\documentclass[journal]{IEEEtran}
\usepackage{cite}
\usepackage{amsmath,amssymb,amsfonts}
\usepackage{algorithmic}
\usepackage{graphicx}
\usepackage{textcomp}
\usepackage{caption}
\usepackage{subfigure}
\ifCLASSINFOpdf
\else
\fi
\begin{document}
%
\title{Program Classification Using Gated Graph Attention Neural Network for Online Programming Services}
%
%
%

\author{Mingming~Lu,~\IEEEmembership{Member,~IEEE,}
        Dingwu~Tan,
        Naixue~Xiong,~\IEEEmembership{Senior Member,~IEEE,}
        Zailiang~Chen,
        and~Haifeng~Li,~\IEEEmembership{Member,~IEEE}
\thanks{M. Lu, D. Tan, and Z. Chen are with the School of Computer Science and Engineering, Central South University,
Changsha, China, 410083, e-mail: (mingminglu@csu.edu.cn, dingwutan@163.com, xxxyczl@csu.edu.cn).}
\thanks{N. Xiong is with the College of Intelligence and Computing, Tianjin University, Tianjin, China, 300350, e-mail:(xiongnaixue@gmail.com).}
\thanks{H. Li is with the School of Geosciences and Info-Physics, Central South University,
Changsha, China, 410083, e-mail: (lihaifeng@csu.edu.cn).}%
}

%
%

\markboth{Journal of \LaTeX\ Class Files,~Vol.~14, No.~8, August~2015}%
{Shell \MakeLowercase{\textit{et al.}}: Bare Demo of IEEEtran.cls for IEEE Journals}
%



\maketitle

\begin{abstract}
The online programing services, such as Github, TopCoder, and EduCoder, have promoted a lot of social interactions among the service users. However, the existing social interactions is rather limited and inefficient due to the rapid increasing of source-code repositories, which is difficult to explore manually. The emergence of source-code mining provides a promising way to analyze those source codes, so that those source codes can be relatively easy to understand and share among those service users. Among all the source-code mining attempts, program classification lays a foundation for various tasks related to source-code understanding, because it is impossible for a machine to understand a computer program if it cannot classify the program correctly. Although numerous machine learning models, such as the Natural Language Processing (NLP) based models and the Abstract Syntax Tree (AST) based models, have been proposed to classify computer programs based on their corresponding source codes, the existing works cannot fully characterize the source codes from the perspective of both the syntax and semantic information. To address this problem, we proposed a Graph Neural Network (GNN) based model, which integrates data flow and function call information to the AST, and applies an improved GNN model to the integrated graph, so as to achieve the state-of-art program classification accuracy. The experiment results have shown that the proposed work can classify programs with accuracy over 97\%.
\end{abstract}


%
\IEEEpeerreviewmaketitle

\section{Introduction}
\label{sec:introduction}
Recently, numerous large-scale online programming services, focusing on programming sharing, exercising, and learning, have received increasing attention. These online services have promoted a lot of social interactions among the service users, such as programers, educators, and IT companies. For example, Github~\cite{b00} motivates programmers to share their codes and promote cooperation among programmers; TopCoder~\cite{b01} provides a platform for software design \& implementation, which inspires IT companies to recruit potential employees from the top coders; EduCoder~\cite{b02} constructs a platform for starters to learn and excercise programming, which encourages the communication between programing educators and programming starters.

However, in the existing online programming services, users have to manually explore the resources that best match their interests. For example, in Github, most cooperated projects rely on a single programer to complete the most significant part of the projects~\cite{b0}, due to the rapid increasing of source-code repositories, which make it hard to match the requirement of the projects and the capability of project contributors; similarly, in TopCoder, it is difficult to match the requests from the IT companies to the programming skills from the programmers; finally, in EduCoder, it is a challenge to suggest the best programming exercises to starters according to their programming levels.  

The emergence of source code mining~\cite{b1} provides a promising way to analyze the source codes submitted by service users. By analyzing those source codes through various machine learning and deep learning techniques, those programming online services can 
help understand and classify the programming projects, and make suggestions for social interactions among service users accordingly. For example, source code mining can classify the shared source codes in Github, TopCoder, and EduCoder, according to the semantic meaning of the codes. Based on the code/program classification, those online programming services can analyze the programming capability of the program authors and recommend the programs to the appropriate service users. Moreover, the program representation learned by program classification can  pave the way to understand the source codes~\cite{b36}.

Inspired by the Natural Language Processing (NLP), at the early stage of source code mining, researchers tended to regard source codes as text sequences and apply NLP models to the source codes to perform various prediction tasks ~\cite{b3,b8,b10,b16,b29}. However, unlike nature languages, which have a relatively fixed vocabulary, the possible class/function/variable names appear in source codes can be unlimited. Therefore, two programs with the same syntax structure and the same semantic meaning, but with different variable names, may be regarded as different programs, as illustrated by the example codes shown in Figure1.

\begin{figure}
\centering
\subfigure[An example code.]{
\label{fig:NLP_compare1} 
\includegraphics[width=0.190\textwidth,height=0.145\textheight]{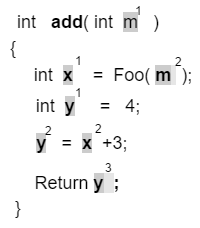}}
\subfigure[The same code with different variable names.]{
\label{fig:NLP_compare2} 
\includegraphics[width=0.195\textwidth,height=0.150\textheight]{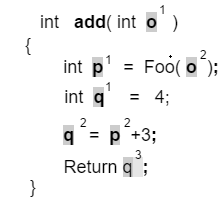}}
\centering
\caption{An example to illustrate the drawback of the NLP-based method.}
\label{fig:NLP_compare} 
\end{figure}

To address the above problem, several works ~\cite{b5,b20,b24,b25,b26,b38} proposed to adopt Abstract Syntax Tree (AST) to characterize the syntax structure information of the source code. The AST corresponding to the program in Figure~\ref{fig:NLP_compare1} is shown in Figure~\ref{fig:AST_Example}, which is also the AST for the code in Figure~\ref{fig:NLP_compare2}. Note that the function/variable names shown in Figure~\ref{fig:AST_Example} is just for illustration, which in fact do not appear in the AST.

Although the AST-based techniques can utilize the syntax information from the source codes, they cannot reflect numerous semantic information embedded in the source codes. Therefore, recent works~\cite{b2,b13} proposed to attach additional data flow edges to the AST and apply the Gated Graph Neural Networks (GGNN) to the extended AST for suggesting variable names. However, these works only consider the fine-grained and local data flows among the variables. To include the coarse-grained and global function-call relations, we proposed to further attach the function-call edges to the extended AST and apply an improve GGNN model, called Gated Graph Attention Neural Networks (GGANN), which integrates the attention mechanism to GGNN, to program classification.

To evaluate the proposed work, we conduct experiments based on the source codes collected from an online judge system, i.e., an online program exercise system,  from three aspects. The experiment results show that the average classification accuracy of the proposed model is 97.8\% and the learned graph nodes representation reflects their semantic information in source codes .

In general, our contributions can be summaried as follows: (1) we propose a program graph, which integrates both data-flow and function-call information into AST to characterize both syntactic and semantic information from the source codes; (2) we improve the GGNN model by introducing the attention mechanism to learn the weight of each individual nodes in the program graph and the aggregated representation of the whole program, and utilize the learned representation to classify the programs; (3) we evaluate the performance of the proposed work through comparative experiments.

The rest of this work is organized as follows. Section~\ref{relatedWork} describes the related works. Section~\ref{preliminaries} introduces the preliminaries concepts and models, which are necessary to apprehend the proposed work. Section~\ref{architecture} presents the proposed architecture, the program graph construcdtion, and the details of the GGANN model. Section~\ref{experiment} provides experimental results and analysis. Section~\ref{conclusion} summarizes this work and discusses about the further work.

\begin{figure}

\centering
\includegraphics[width=0.25\textwidth]{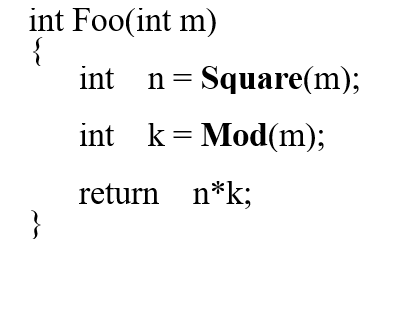}
\caption{The function called by the code in Figure~\ref{fig:NLP_compare}.}
\label{fig:NLP_functions_called} 
\end{figure}


\section{Related Work}\label{relatedWork}

In the earlier stage of source code mining, most of the existing works focused on the NLP-based techniques, which process source codes as tokens~\cite{b3,b8,b38} or APIs~\cite{b10,b16} sequences. The related tasks are to predict sequences composition, translation between natural language and code, code completion, and etc. Hindle et al.~\cite{b8} believed that source code has certain statistical properties, similar to natural languages. Thus, they used n-gram, a popular statistical model in NLP,  to model source codes and illustrated that their model can enhance the Eclipse's ability to complete Java code. Bhoopchand et al.~\cite{b4} used the Pointer Network model for end-to-end training of source code to achieve IDE code completion task. Gu et al.~\cite{b10} adopted  Encoder-Decoder based RNN to obtain API usage sequences from API-related natural language queries.

To extract syntactic and semantic information from source code, Mou et al.~\cite{b24,b25,b26} proposed a Tree-Based Convolutional Neural Network (TBCNN) to transform the AST into a distributed vector that preserves semantic information. Bui et al.~\cite{b5} utilized double TBCNNs to achieve source code translation between different program languages.

The popularity of Graph Neural Networks (GNN)~\cite{b7,b13,b17} inspired researchers to applying GNN to source code mining~\cite{b2,b13}. The core idea is to represent the program as a simple graph and then feed the graph as an input into GGNN to obtain the distributed node representation for each node in the graph. Finally, the distributed representation is used to detect error variables or predict variable names.

The most related work to our work is the work of Allamanis~\cite{b2}, which applied GGNN to the program graph integrating AST and data-flow edges, in order to predict variable names. However, this work did not take into accounts the function-call relations embedded in the programs, the dynamic distributed vector representations for edges in the program graph, the influence of different adjacent nodes through attention mechanism, and the aggregation of global information through attetion mechanism. To the best of our knowledge, our work is the first work that applies the GNN-based model to program classification task. 


\section{Preliminaries}\label{preliminaries}

This section mainly introduces some basic concepts used in this paper, including graph, graph neural network, gated graph neural network.

\subsection{The Graph Model}

\begin{figure}
\centering
\subfigure[Example]{
\label{fig:codey3z} 
\includegraphics[scale=0.5]{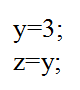}}
\hspace{.1in}
\subfigure[Data flow graph]{
\label{fig:dfgy3z} 
\includegraphics[scale=0.4]{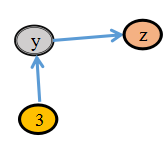}}
\centering
\hspace{.1in}
\subfigure[Connection matrix]{
\label{fig:connectiony3z} 
\includegraphics[scale=0.4]{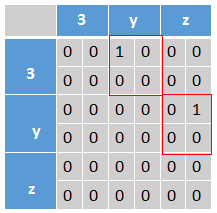}}
\centering
\caption{An example to illustrate the connection matrix and the propagation matrix.}
\label{fig:y3z} 
\end{figure}

The graph model used in this paper is a directed graph $G=(V,E)$, among which $V$ represents the node set of size $|V|$ and depicts the directed edge set of size $|E|$. A node in $V$ is identified by a number and a directed edge in $E$ is expressed by $e_{ij}$, representing a directed edge from node $i$ to node $j$ . An edge type set $L_{K}=\{l_1,l_2,......,l_k\}$ is used to represent different types of edges in $E$. The connection relations among nodes in graph $G$ is represented by a connection matrix $A \in R^{|V|\times2|V|}$, where the element $A_{ij}$ in row $i$ and column $j$ is a $d \times d$ matrix, where $d$ represents the feature dimension of a node. $A_{ij}$ is the propagation matrix of edge $e_{ij}$, which represents the information propagation rule from node $i$ to node $j$. For example, Figure~\ref{fig:connectiony3z} illustrates the connection matrix corresponding to the data flow graph shown in Figure~\ref{fig:dfgy3z}, which in turn is generated by the example code shown in Figure~\ref{fig:codey3z}. Two red rectangular matrices are propagating matrices corresponding to $e_{3y}$ and $e_{yz}$, respectively.

In the assignment statements shown in Figure~\ref{fig:codey3z}, we are concerned about whether the immediate number $3$ can be passed to the variable $z$, as shown in Figure~\ref{fig:dfgy3z}. Therefore, node $3$ and node $z$ can be regarded as the source and target nodes, respectively, and their feature vectors can be initialized with $h_{3}^{0}=[1,0]$ and $h_{z}^{0}=[0,1]$, respectively, where the first dimension and and the second dimension of $h_{3}^{0}$ and $h_{z}^{0}$ represent the source node and the target node, respectively. Since node $y$ is neither a source node nor a target node, its feature vector is initialized as $h_y^0 = [0,0]$.

Propagation matrix $A_{ij}$ determines the amount of information associated with each dimension of node $i$ propagated to each dimension of node $j$, where 0 means no information propagation, while 1 means complete information propagation. For example, $A_{3y}$ in Figure~\ref{fig:connectiony3z} indicates that node $3$ only passes its first dimension information to the first dimension of node $y$. Thus, the propagation of information from node $3$ to node $y$ through the multiplication between vector $h_3$ and matrix $A_{3y}$ is $h_y=[1,0]$, indicating that the data has not be transmitted to the target node, $z$. However, $A_{yz}$ in Figure~\ref{fig:connectiony3z} denotes the first dimension information of node $y$ will be transmitted to the second dimension of node $z$. Therefore, the result vector $h_z=[0,1]$ obtained through the multiplication between vector $h_y$ and matrix $A_{yz}$ implies that the data has been transferred to the target node $z$.

\subsection{Graph Neural Network}

In Graph Neural Network (GNN)~\cite{b21,b30}, each node updates its hidden state information by aggregating its neighbors' hidden states and its own state information at the previous time step, so as to predict the node attributes or the attributes of the whole graph. In order to aggregate information, $T$ iterations of state information propagation will be applied to each individual GNN node. For each node $i$, its state information is initialized as a vector $h_i^{(0)} \in R^d$. In the t-th iteration of the information propagation, node $i$ gathers its neighbors' state information and aggregates them as $m_t^{(t)}$, as shown in Formula~\eqref{GNN_eq_1}, where $N_i$ represents node $i$ neighbor set. Then, the aggregated neighbor information is combined with node $i$ 's previous state information $h_i^{(t-1)}$ to form the new state information $h_i^{(t)}$ through a neural network, called $nn$, as shown in Formula~\eqref{GNN_eq_2}. It is worthy to note that the parameters of the propagation matrix  $A_{ij}$ in Formula~\eqref{GNN_eq_1} are learned and shared for the same type of edges.

\begin{equation}
 \label{GNN_eq_1}
 m_i^{(t)}= \sum_{j \in N_i}A_{ij} \cdot h_j^{(t-1)}
\end{equation}

\begin{equation}
\label{GNN_eq_2}
 h_i^{(t)} = nn(m_t^{(t)},h_i^{(t-1)})
 \end{equation}

\subsection{Gated Graph Neural Network}

Gated Graph Neural Network (GGNN) is actually an extension of GNN by replacing function $f$ in Formula~\eqref{GNN_eq_2} with the GRU function shown in Formula~\eqref{GGNN_eq_1}. The GRU function enables each node to memorize history information so that the temporal sequential relations among $h_i^{(t)}$, $h_i^{(t-1)}$ and $m_i^{(t)}$ can be taken into account.

\begin{equation}
\label{GGNN_eq_1}
 h_i^{(t)} = GRU(h_i^{(t-1)},m_i^{(t)})
 \end{equation}

\begin{figure*}[ht]
    	\centering
    	\includegraphics [width=1\textwidth]{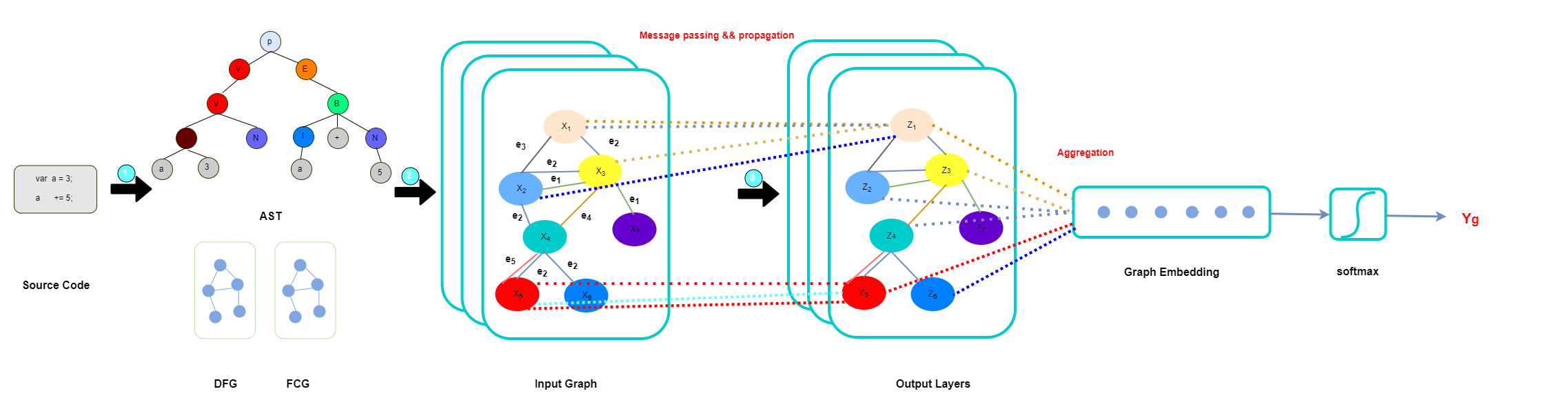}
    	\caption{The GGANN architecture.}
    	\label{fig:modelArchitecture}
\end{figure*}

\section{The GGANN Architecture}\label{architecture}

This section mainly presents the architecture of the proposed work, as shown in Figure~\ref{fig:modelArchitecture}, which consists of two main components: the construction of the program graph and the application of the proposed GGANN model to the program graph. The first component converts the source code of a program to a program graph, while the second one learns the distributed representation of the program graph and utilizes the learned representation to classify the program. In the following, we will present each component individually.

\subsection{The Program Graph Construction}
As shown in Figure ~\ref{fig:modelArchitecture}, the program graph construction itself actually consists of two main steps: 1) the construction of the Abstract Syntax Tree (AST), the Function Call Graph (FCG), and the Data Flow Graph (DFG) individually by parsing the source code; 2) the construction of the program graph by integrating the AST, FCG, and DFG. In the following, we will present the four types of graphs individually.

\subsubsection{Abstract Syntax Tree (AST)}

\begin{figure}[t]
    	\centering
    	\includegraphics [width=0.5\textwidth]{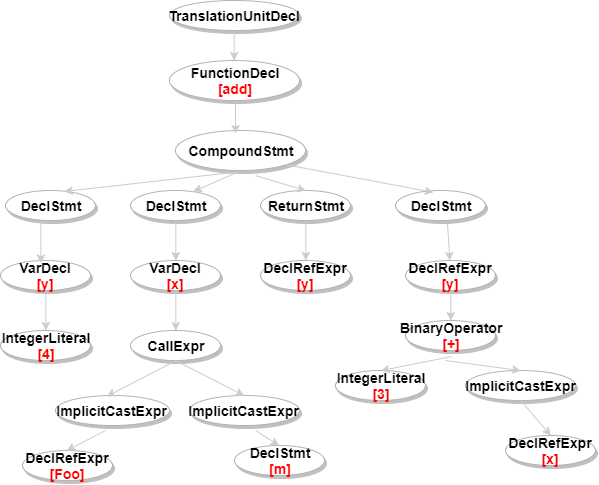}
    	\caption{The AST generated from the program segment shown in Figure~\ref{fig:NLP_compare}.}
    	 \label{fig:AST_Example}
\end{figure}

AST~\cite{b22} is an intermediate representation of a program generated through the program syntax and semantic analysis. It uses context-free grammar parsing rules\footnote{https://en.wikipedia.org/wiki/Context-free\_grammar} to construct a tree structure that characterizes the grammar structure and execution order of a program. Thus, an AST can equivalently represent the syntax structure of a program. Each node within the AST represents an element in the program. To illustrate AST, we use the program segment shown in Figure~\ref{fig:NLP_compare1} to generate the corresponding AST as shown in Figure~\ref{fig:AST_Example}, where all the elements in the program shown in Figure~\ref{fig:NLP_compare1} are represented as nodes in the AST. To illustrate the correspondence between the program and the AST, we attach the variable/function/operator/literal elements in the program to the corresponding nodes in the AST. For example, in Figure~\ref{fig:AST_Example}, the FuctionDecl node is attached with `[add]', which denotes that the node represents the add function declaration. Therefore, the subtree with the FunctionDecl node as the root node represents the body of the add function. Similarly, the VarDecl node represents a variable declaration, and the IntegerLiteral node indicates that an integer literal is declared.

As the parse tree of a program, an AST mainly covers the following basic structures: selection structure (such as IF and SWITCH), loop structure (such as WHILE and FOR), sequential structure (such as expression and assignment). Therefore, AST, as an intermediate representation of a program, effectively preserves the syntax of the associated programming language and the context of the program runtime.

\subsubsection{Function Call Graph (FCG)}
\begin{figure}[t]
    	\centering
    	\includegraphics [width=0.3\textwidth]{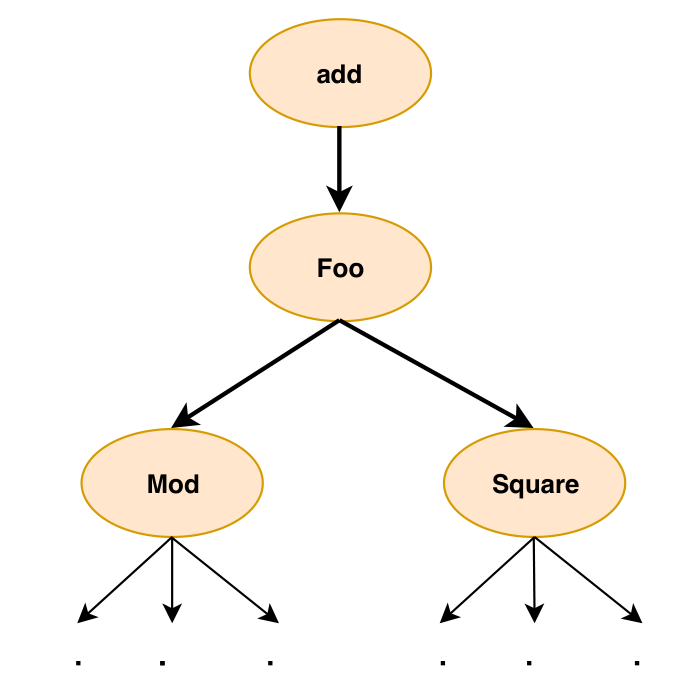}
    	\caption{The FCG example}
    	    	\label{FCG}
\end{figure}

FCG~\cite{b34} models the function semantic information in a program. Each node in FCG represents a function, while each edge in FCG denotes the function-call relation between two functions. To illustrate FCG, from the code fragment in Figure~\ref{fig:NLP_compare1} and Figure~\ref{fig:NLP_functions_called}, we generate the corresponding FCG, as shown in Figure~\ref{FCG}, where the name of a node corresponds to the name of the corresponding function in Figure~\ref{fig:NLP_compare1} and Figure~\ref{fig:NLP_functions_called}, and the existence of an edge is determined by the function-call  relation between two function nodes.

\subsubsection{Data Flow Graph (DFG)}

To capture the data flow and control flow information associated with the program runtime, we use DFG~\cite{b39} to characterize this unique semantic information. A node in DFG can be an entity, such as a variable, a operator, a structure identifier, and etc, while an edge in DFG represents the data transfer between two entities. DFG can be used to characterize the logic and functionality of a program.

To illustrate DFG, based on the code shown in Figure~\ref{fig:NLP_compare}, we generate the corresponding DFG, as shown in Figure~\ref{DFG}, where a superscript associated with a variable node denotes the status of the corresponding variable. For example, in Figure~\ref{DFG}, variable $y$ has 3 status, namely, $y^1$, $y^2$, and $y^3$, respectively. The reason to introduce the superscript to describe various status lies in that the value stored in a variable may change during the process of program execution. Note that there exist numerous situations, where a variable may be modified or modify a variable value depend on a particular variable status. For example, variable $y$ in the program fragment shown in Figure~\ref{fig:NLP_compare1} has three status, which correspond to two assignment statments, and one return operation. 

To characterize the above relations, we adopt 5 types of DFG edges, namely, ``LastUse'', ``Compute'', ``Formal'', ``Return", and ``Operand'', which are illustrated through 5 different colors in Figure~\ref{DFG}. A ``LastUse'' edge characterizes the relation between two different status of the same variable. For example, consider the program shown in Figure~\ref{fig:NLP_compare1}, where the dark green dotted edges denote the ``LastUse'' edges.  The edge from $x^2$ to $x^1$ is an ``LastUse'' edge  because the value of variable $x$ in statment ``$y=x+3$''  directly depends on the value of $x$ in statement ``$int$  $x = FOO(m)$''. 

A ``Compute'' edge represents the relation, where a variable is computed from the otherf variable or a function, as shown in the purple dotted edges in Figure~\ref{DFG}. For example, in statment ``$int$ $x^1=Foo(m^2)$'', $x^1$ is computed from function $Foo$ and variable $m^2$. A ``Return'' edge represents the relation from the function return value to the function name, which can be illustrated through the red dotted edge from $y^3$ to $add$ in Figure~\ref{DFG}. An ``Operand'' edge denotes the relation from the operator to the operand in an expression, which can be illustrated through the light green dotted edge from $+$ to $y^2$ in Figure~\ref{DFG}. 

A ``Formal'' edge describes the relation from the formal argument of a caller function to the formal argument of a called function,  or the relation from the actual argument in a called function to the formal argument in the caller function, which can be illustrated through the orange dotted edges in Figure~\ref{DFG}. Note that the ``Formal'' edge can also describe the argument transfer among functions. For example, in Figure~\ref{DFG}, node $m^{Foo}$ represents the formal parameter $m$ when function `Foo' is declared, and $m^1$ denotes the actual parameter passed to `Foo' when it is called by function `add'. If the called function has multiple arguments, it is necessary to connect the actual parameters and the formal parameters through multiple ``Formal'' edges correspondingly.


\subsubsection{The Integrated FDA Graph}

Through comparing AST, FCG, and DFG, it can be observed that each of them can only represents partial information of a program. The AST basically contains the grammatical structure information of a program and a small amount of semantic information, while an executable program has richer semantic information at runtime. FCG and DFG describe the semantic runtime information of a program from the perspective of code blocks (such as, functions, methods, and etc.) and code identifiers (such as, variables, operators, and etc.), respectively.

Therefore, it is desirable to integrate these three code representations into a new program graph. Since the AST contains both function nodes and variable/operator nodes, it is convenient to add a FCG edge or a DFG edge to the AST if there exists a function-call relation or data-flow relation between two AST nodes. Through inserting the FCG or DFG edges accordingly, the integrated program graph can not only represent the syntax information but also the function-call and data-flow semantic information. We call the integrated program graph as the FDA graph, which embeds not only the basic grammatical structure information, but also the runtime information. Therefore, the FDA graph can effectively reduce the information loss due to the conversion of a program to an intermediate representation.

\begin{figure}[t]
    	\centering
    	\includegraphics [width=0.35\textwidth]{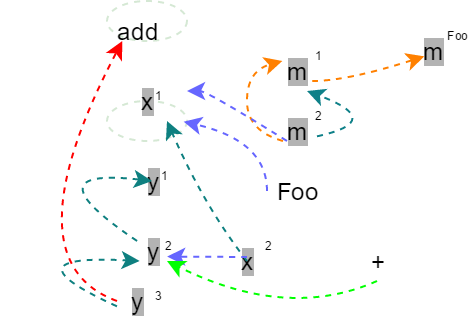}
    	\caption{The DFG example}
    	    	\label{DFG}
\end{figure}

\begin{figure}[htb]
    	\centering
    	\includegraphics[width=0.5\textwidth]{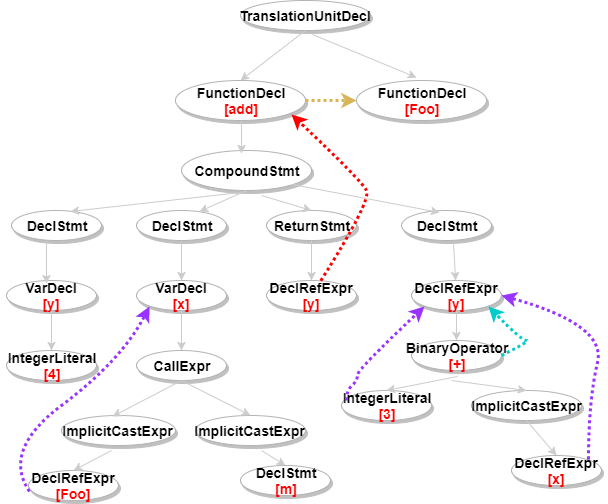}
    	\caption{The FDA example.}
    	\label{fig:FDA_example}

\end{figure}

To illustrate the integration of AST, FCG, and DFG, we added the FCG edges and the DFG edges to the AST in Figure~\ref{fig:AST_Example}. The integrated FDA graph is shown in Figure~\ref{fig:FDA_example}, where the yellow dotted edge represents the FCG edge and the colors of the other dotted edge denotes the same types of DFG edges shown in Figure~\ref{DFG}.

The integrated FDA graph has 7 edge types in total, including the original edge in the AST, the 5 types of edges introduced by the DFG, and the function-call edge. The enriched FDA graph can speed up the information propagation on the proposed GGANN model, which will be presented as follows.

\subsection{The GGANN model}

In GGNN, the node connection and information propagation rule are defined by the connection matrix $A$, where the propagation matrix are the same for the same type of edges and also will not change over different period of time. However, the same information propagation path may have different semantic information for different contexts. For example, consider a conditional selection code block in a program, where there exists program fragment right after both the ``if'' and ``else'' branches. The execution of any of these two branches represent different contexts for the followed program fragment. Therefore, it is necessary to dynamically learn the parameters of the propagation matrix according to the context and state information. For this reason, we first introduce d-dimensional vector $h_{ij}^{'}$ to represent the hidden state of the directed edge $e_{ij}$, which reflects the dynamic variation of the edge state. $h_{ij}^{'}$ can be implemented by a multi-layer neural network $U_e$, the input of which includes both node i's and node j's hidden states, as well as the previous edge state information, as shown in Formula~\eqref{GGANN_eq_1}.

\begin{equation}
 \label{GGANN_eq_1}
 h_{ij}^{'(t)} = U_e(h_{ij}^{'(t-1)},h_i^{(t-1)},h_j^{(t-1)})
\end{equation}

Based on this, we can replace the propagation matrix $A_{ij}$ in Formula~\eqref{GNN_eq_1} with a propagation matrix function, $A(h_{ij}^{'}):R^d \rightarrow R^d \times R^d$, which learns the propagation matrix dynamically according to the dynamic information of the hidden state $h_{ij}^{'}$. $A(h_{ij}^{'})$ can also be realized by a multi-layer neural network, which, at time $t$, maps the hidden state $h_{ij}^{'(t)}$ to the propagation matrix.

In addition, for each node, its neighbor nodes may have different importance in terms of the contribution to update the node hidden state. However, as shown in Formula~\eqref{GNN_eq_1}, both GNN and GGNN simply sum the neighbors' state information of a node. Therefore, we introduce the information strength $\alpha_{ij}$ to quantify the information contribution from node $j$ to node $i$. $\alpha_{ij}$ can be implemented by the attention mechanism $a:R^d \times R^d \rightarrow R$, which compares the similarity between node $i$ and its neighbor node $j$, and the similarity is normalized through the activation function, softmax, as shown in Formula~\eqref{GGANN_eq_2}:

\begin{equation}
 \label{GGANN_eq_2}
 {\alpha}_{ij}= softmax(a(h_i^{(t-1)},h_j^{(t-1)}))
\end{equation}

The parameters of the neural network $a$ are shared for the edges with the same type and the same direction.

In a summary, the node information aggregation adopted in GNN and GGNN, as shown in Formula~\eqref{GNN_eq_1}, can be replaced by the  
Formula~\eqref{GGANN_eq_3} in GGANN.

\begin{equation}
 \label{GGANN_eq_3}
 m_i^{(t)} = \sum_{j \in N_i}{\alpha_{ij}A(h_{ij}^{'(t)})h_j^{(t-1)}}
\end{equation}

Finally, because the program classification task is to classify a program through the associated FDA graph, the vector representation of the FDA graph should be obtained after the calculation of each node's embedding vector, i.e., its hidden state. Since different nodes in the FDA graph may contribute differently to classify the program, we adopt the soft attention mechanism to compute the weight of each node to the FDA graph.
The soft attention mechanism involves the similarity comparison between each node and the entire graph. 

However, the vector representation for the FDA graph is unknown before the calculation of a node's contribution. To address this, we use the updated hidden state of each node after T iterations of information propagation to approximate the vector representation of the entire FDA graph. The underlying reason lies in that each node contains the information of its T-hop neighbor nodes after T iterations, which can well approximate the whole graph with sufficient large T.

Besides that, each node's initial information usually uses the node's input feature vector. Therefore, the weight of each node in the FDA graph can be replaced by $f(h_i^{(T)},x_i)$, where $f$ represents a neural network to implement soft attention mechanism, and the output of $f$ is the similarity (a scalar) between the node and the entire FDA graph. Based on this, the weighted embedding vector of a node can be calculated by multiplying each dimension of the node embedding vector $g(h_i^{(T)})$ with the weight of the node.

Thus, the embedding vector of the entire FDA graph can be computed by summing the weighted embedding vector of each node. Once the embedding vector of the FDA graph is obtained, the category of the program associated with the FDA graph can be predicted by applying the softmax activation function to the embedding vector, as shown in Formula~\eqref{GGANN_eq_4}.

\begin{equation}
 \label{GGANN_eq_4}
 h_G = softmax(  \sum_{i \in V}  f(h_i^{(T)},x_i) \bigodot g(h_i^{(T)}))
\end{equation}

\section{Experimental Evaluation}\label{experiment}
To evaluate the effectiveness of the proposed GGANN model and the FDA graph, we not only evaluate the model from the perspective of program classification accuracy, but also analyze what the proposed model learns from the perspective of the semantic information implied by the learned embedding vectors. To understand the learning effect of the model intuitively, some experimental results are presented visually.

\subsection{The Dataset Description}

The experimental data are collected from an Online Judge (OJ) programming exercise system, where administrators issue numerous programming tasks, to each of which students submit their coded solutions in the form of source-code files. To evaluate the correctness of the submitted codes, the OJ system will compile and execute the codes and compares the execution results with the standard output. Therefore, the data set is mainly composed of two parts: the programming tasks (represented by question number) and the corresponding source codes submitted by students.

To exclude the influence of programming languages, we consider only the source codes written in C++. Meanwhile, to ensure that there are sufficient source codes for each programming task, we filter the data so that only the programming tasks with more than 350 submitted codes will be considered for program classification. In total, 30 programming tasks meet this filtering requirement. Since the selected programming tasks are more common than the filtered tasks, we call them the common data subset. 

Among these common tasks, there exist a few tasks, which are similar in terms of task description and coding implementation. These similar tasks can be utilized to evaluate the robustness and effectiveness the proposed GGANN model and FDA graph. Thus, 10 programming tasks with similar descriptions are selected from the common data sets, and divided into two groups according to their similarity (5 tasks in each group).

We use clang\footnote{http://clang.llvm.org/}, an open-sourced dependency library, to parse each C++ program into an AST, which is further processed to generate the FDA graph by adding data-flow can function-call edges. The generate FDA data set is divided into the training set, the validation set and the test set with the ratio being 3:1:1. Table 1 lists the statistical information of the training, validation and test sets, where the common task and the similar task datasets are represented by Common and Similarity, respectively.

 \begin{table}[htbp]
    \center

    \caption{ Data set statistics}

  \setlength{\tabcolsep}{0.7mm}{

    \begin{tabular}{c|cccccc}
    \hline
    & &Graphs&Nodes&Edges&Classes&Edge Types\\
    \hline
   Train&Common &35643  &8456506	&10895549   &30  &7 \\
   &Similarity  &7123  &1408234	    &1884146   &10  &7 \\
   \hline
    \hline
   Valid&Common &11297  &2838685	&3994853   &30  &7 \\
   &Similarity  &2411   &461078	    &649538    &10  &7 \\
   \hline
   Test&Common   &11366   &3061059   &4259252   &30    &7 \\
   &Similarity   &2583   &458344	 &650955    &10    &7 \\
    \hline
    \end{tabular}}\label{table:dataStatistics}
    \end{table}

\subsection{The Experiment Set-up}

To train the model, the optimization algorithm adopted the ADAM optimizer's Stochastic Gradient Descent (SGD) algorithm~\cite{b18}, the loss function is to minimize the cross entropy, the batch size is set as 10000, and the number of epoch is set as 3000. The model parameters are initialized with the Glorot initialization method (set to 0.0001)~\cite{b33}. To avoid overfitting, we adopted the $L_2$ regulization term with the initial value $\lambda=0.0005$, and applied dropout~\cite{b31} to the input of each layer with probability $\rho=0.6$. We also used a linear learning-rate decay to adjust the learning rate from the initial learning rate $l$ to the final learning rate $l * F$ , with the attenuation coefficient $F$ in the range [0, 1].

\begin{figure*}
\centering

\subfigure[The training speed with different d values]{
\label{fig:dimenSpeed}
\includegraphics[scale=0.5]{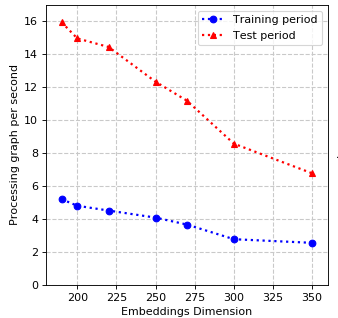}}
\subfigure[The loss with different d values]{
\label{fig:dimenLoss}
\includegraphics[scale=0.5]{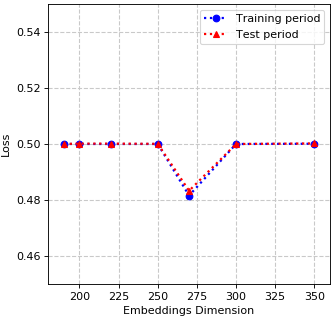}}
\centering
\caption{The determination of the d values}
\label{fig:determineDimension} 
\end{figure*}

In our experiment, the GGANN network has 5 hidden layers representing 5 iterations of information propagation. The validation set is used to choose the model hyper parameters and determine the opportunity for early stopping, while the test set is used to calculate the classification accuracy. For example, the number of hidden layer neurons, i.e., the dimension of node features d, used in the experiment, is a hyper parameter. d can be determined by the following two factors: (1) the convergence speed of the model, (2) the loss value. To determine the appropriate value of d, several experiments have been conducted. The experimental results are shown in Figures~\ref{fig:dimenSpeed} and \ref{fig:dimenLoss}. Figure~\ref{fig:dimenSpeed} shows the variation of the training/testing speed in terms of the number of graphs processed per second along with the change of the d value, while Figure~\ref{fig:dimenLoss} shows the variation of the loss value along with the change of d value for both training and test sets. As can be seen from both Figures~\ref{fig:dimenSpeed} and \ref{fig:dimenLoss}, when d is set to 270, the loss value of the model is relatively small, and the training speed is relatively fast. Therefore, in the subsequent experiments, we set the hidden layer vector dimension d = 270.

\subsection{The Experiment Results}

This subsection mainly analyzes the program classification performance associated with the proposed GGANN model and the FDA graph, as well as what the embedding vectors learned, through experimental evaluation. We first analyze the program classification performance.

\subsubsection{Program Classifying}

The experiments for program classification are used to verify whether the proposed work can successfully learn grammatical structure and semantic information from source codes, i.e., whether the submitted codes implemented by different students for the same programming task can be classified into the same category, while the codes for different programming tasks with similar grammatical structure can be effectively distinguished.

In the experiments, we consider binary classification, which determines whether a given program belongs to the associated programming task or not. For each program, the corresponding programing task number is used as its label. The baseline is the Tree Based Convolution Neural Network (TBCNN), which, as far as we know, is the state-of-art work for program classification. In addition, since the proposed GGANN model is an extension of the GGNN model, we also adopt GGNN as the baseline model. It is worthy to note that TBCNN cannot apply to the FDA graph, because it is designed for AST only.

Table~\ref{table:similarityClassification} shows the experiment results in terms of the average, minimum, and maximum program classification accuracy for the GGANN, GGNN and TBCNN models on the similar datasets. The experimental results show that the classification accuracy of the TBCNN model is significantly lower than those of the GGANN and GGNN models, and the GGANN model is also significantly improved compared with the GGNN model. From the experiment results, the following observation can be concluded. On one hand, TBCNN relies too much on AST, which makes it difficult for TBCNN to distinguish similar tasks with close static structures but different runtime semantic information. On the other hand, the edge embedding representation and the attention mechanism adopted by the GGANN model effectively improve the GGNN model.

\begin{table}[htbp]
    \center

    \caption{The program classification accuracy of different models on the similar programming tasks.}

  \setlength{\tabcolsep}{0.7mm}{

    \begin{tabular}{cccc}
    \hline
    	Accuracy&Average &	Minimum &Maximum \\
    \hline
Similarity(GGANN)&	\textbf{93.9\%}&	\textbf{88.7\%}&	\textbf{97.7\%} \\
Similarity(GGNN)	&90.3\%&	84.2\%&	94.8\%  \\
Similarity(TBCNN) &85.7\%& 83.2\%   &87.9\%\\
    \hline
    \end{tabular}}\label{table:similarityClassification}
    \end{table}

To further clarify the origin of the performance improvement, i.e., whether the improvement comes from the model advantage of GGANN over TBCNN and GGNN or the information advantages of FDA over AST, we further compare the performance of the three models to AST and the performance of GGANN and GGNN to FDA for the common task data set. The experimental results are shown in Table~\ref{table:threeModelCompare}. Compared with the results in Table~\ref{table:similarityClassification}, it can be concluded that, for all the three models, the classification accuracy on the common tasks is better than those on the similar tasks. 

Furthermore, it can be inferred that, in contrast to GGNN and GGANN, the performance improvement of TBCNN for similar tasks over common tasks is about 9\%, which is significantly higher than those of GGNN and GGANN. This indirectly illustrates that the graph network models (GGNN and GGANN) are more robust, because the performanc of the graph models reduce much less than TBCNN when it deals with similar tasks. 

Moreover, from Table~\ref{table:threeModelCompare}, by observing the accuracy of the three models on AST, it can be inferred that the graph network model alone can significantly improve the classification accuracy. This can illustrate the contribution of the graph models.

 \begin{table}[htbp]
    \center
    \caption{The accuracy of the three models associate with AST and FDA for the common data set}
    \setlength{\tabcolsep}{1mm}{

    \begin{tabular}{cccccc}
    \hline
   && &	TBCNN&	GGNN	&GGANN \\

    \hline

&	&Average&	0.941&	0.964	&\textbf{0.971}
\\
&AST&   Minimum	&0.926&	0.935	&\textbf{0.946} \\
&&Maximum	&0.952	&0.976	&\textbf{0.981}\\
Accuracy&&Average	&-	&0.968	&\textbf{0.978}\\

&FDA&	Minimum	&-	&0.943	&\textbf{0.951}
\\
&&Maximum&	-	&0.978	&\textbf{0.983}\\

    \hline
    \end{tabular}}\label{table:threeModelCompare}
    \end{table}


\begin{figure*}
\centering
\subfigure[Similar tasks (validation set)]{
\label{fig:similarityAccValid}
\includegraphics[scale=0.6]{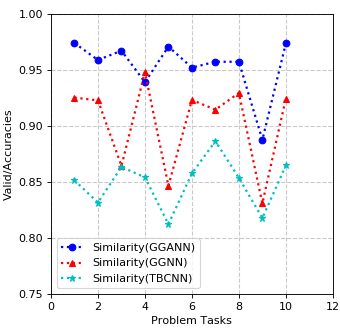}}
\subfigure[Similar tasks(test set)]{
\label{fig:similarityAccTest}
\includegraphics[scale=0.6]{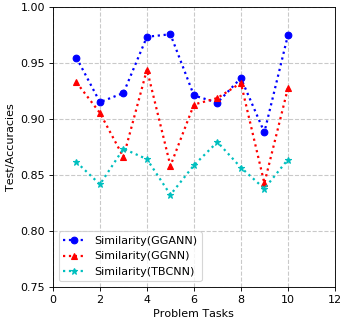}}

\centering
\subfigure[Common tasks (validation set )]{
\label{fig:commonAccValid}
\includegraphics[scale=0.6]{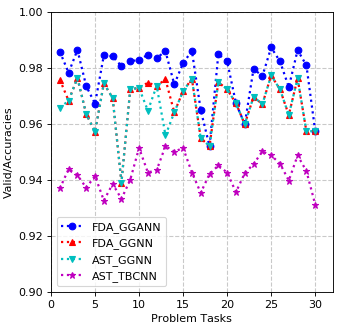}}
\centering
\subfigure[Common tasks (test set )]{
\label{fig:commonAccTest}
\includegraphics[scale=0.6]{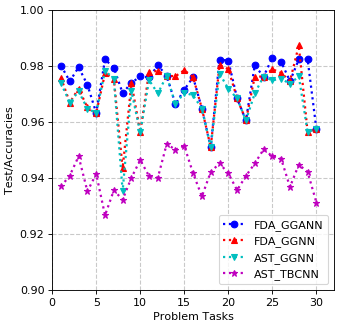}}
\caption{ The classification accuracy of similar/common programming tasks for the combination of models and program intermediate representations }
\label{fig:classificationAccuracy} 
\end{figure*}
However, the experimental results shown in Table~\ref{table:threeModelCompare} also imply that the performance improvement of the graph models on the FDA graph over AST is not significant. In order to further analyze this phenomenon, we compare the classification accuracy of the three models on each programming task of the common and similar task sets, respectively, as shown in Figure~\ref{fig:classificationAccuracy}. 

Figure~\ref{fig:similarityAccValid} and Figure~\ref{fig:similarityAccTest} visualize the classification accuracy of the three models associated with each similar programming task on the validation set and the test set, respectively. From the experimental results, the following can be observed:  (1) the classification accuracy of the GGANN model is slightly higher than that of the GGNN model in almost every programming task; (2) the accuracy variance of GGANN in different tasks is relatively small; (3) both classification accuracies of the two graph models are higher than that of the TBCNN model. 

Figure~\ref{fig:commonAccValid} and \ref{fig:commonAccTest} visualize the classification accuracy from the application of the three models on the two program intermediate representations (FDA or AST) associated with each common programming task on the validation set and test set, respectively. Besides the application of GGANN on FDA (the proposed work) and the application of TBCNN on AST (the state-of-art method), we also consider the application of GGNN on AST and FDA. The application of GGNN on AST can be used to compare the state-of-art method, so as to identify the advantage of GGNN over TBCNN. The application of GGNN on FDA can be used to compare with the application of GGNN on AST and the proposed work, in order to identify the advantage of FDA over AST and the advantage of GGANN over GGNN, respectively. 

From the experimental results shown in Figure~\ref{fig:commonAccValid} and \ref{fig:commonAccTest}, it can be observed that, although the application of the two graph models on FDA perform better than the application of TBCNN on AST, the contribution of the FDA is not significant, because the application of the GGNN model on FDA is just slightly better than that on AST in terms of classification accuracy.

\begin{figure}[htb]
    	\centering
    	\includegraphics[scale=0.5]{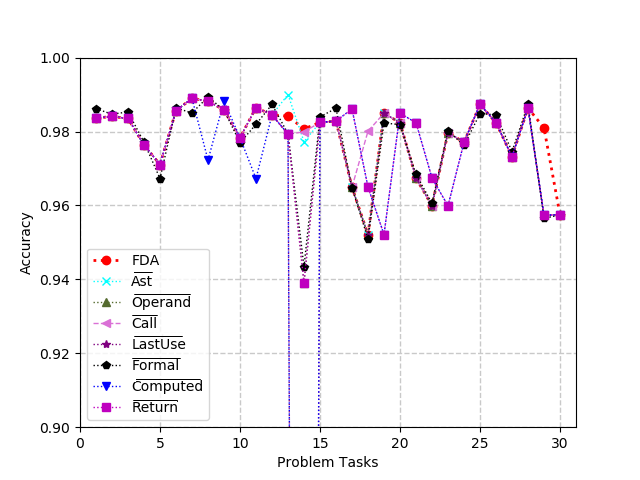}
    	\caption{The contribution comparison among different edge types}
    	\label{fig:deleteEdge}

\end{figure}

\begin{figure}[htb]
    	\centering
    	\includegraphics[scale=0.65]{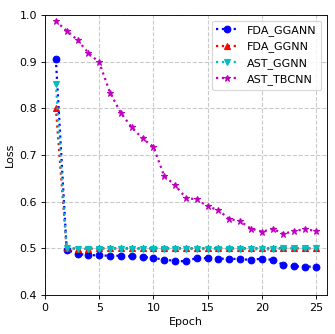}
    	\caption{The variation of loss values for the four models along with the iterations.}
    	\label{fig:loss}
\end{figure}

To evaluate whether the data-flow and function-call edges added to the AST is indeed useful, we remove each of the seven edge types from the FDA graph and apply the GGANN model to the seven modified FDA graphs, respectively, so as to observe the contribution of each edge type to the program classification accuracy. 

The experiment results are visualized in Figure~\ref{fig:deleteEdge}, where $overline{edgeType}$ denotes the modified FDA graph with a given edge type removed. It can be inferred from the experiment results that no edge type contributes significantly more than other edge types except for a few program tasks. Thus, it can be concluded that there exist information redundancy among the seven edge types in most cases, so that the removal of any edge type will not have significant impact on the program classification accuracy. 

However, for a few program tasks, certain edge types contribute much more than the other edge types, i.e., the removal of those edge types can significantly reduce or improve the program classification accuracy, as shown in Table~\ref{table:deleteEdege}, where $\overline{edgeType}$ denotes the modified FDA graph with a given edge type removed. Therefore, this experiment illustrates that each edge type of the FDA graph might be complementary to each other for the program classification, even though they are redundant in most cases.

\begin{table*}[htbp]
    \center
    \caption{The program tasks that can be impacted by certain edge types}
    \setlength{\tabcolsep}{0.6mm}{

    \begin{tabular}{cccccccccc}
    \hline
   & &FDA & $\overline{AST}$ & $\overline{Operand}$ &$\overline{LastUse}$ &$\overline{Compute}$ &$\overline{Return}$ &$\overline{Formal}$ &$\overline{Call}$ \\
    \hline
    &Task13 &0.984 &\textbf{0.990} &0.979 &0.979 &0.979 &0.979 &0.979 &0.979 \\
    \hline
    &Task14 &0.981 &0.977 &0.939 &\textbf{0.061} &\textbf{0.056} &0.939 &0.943 &0.979\\
    \hline
    \end{tabular}}\label{table:deleteEdege}
\end{table*}


In addition to classification accuracy, we also evaluate the impact of the proposed GGANN model and FDA through the loss values. Figures~\ref{fig:loss} show the experimental results, which compare the convergence trend of the loss values associated with the applications of GGANN/GGNN/TBCNN on FDA/AST, respectively, along with the number of iteration epochs. 

From the experiment results, it can be observed that not only the final loss values of the two graph models are smaller, but also they converge faster than TBCNN, even if only the AST edges are used. The reason that graph network model converges faster might lie in that the graph models have stronger constraints on the AST nodes than TBCNN. More specifically, in TBCNN, the convolution operation enforces a one-way information propagation from child nodes to parent nodes, while, in the graph models, for each node, each iteration involves two-way information propagations among all neighbor nodes, which will gradually spread to the whole network. This strong constraint on node relations also enables the stronger learning capability of the graph models than that of TBCNN, as illustrated in Figure~\ref{fig:loss}, where the loss values of GGNN and GGANN almost converge at the beginning. Finally, from Figure~\ref{fig:loss}, it can also be concluded that GGANN is better than GGNN in terms of convergence speed of loss values.


\subsubsection{The Semantic Analysis of the Learned Representation}

\begin{figure}[htb]
    	\centering
    	\includegraphics [width=0.5\textwidth]{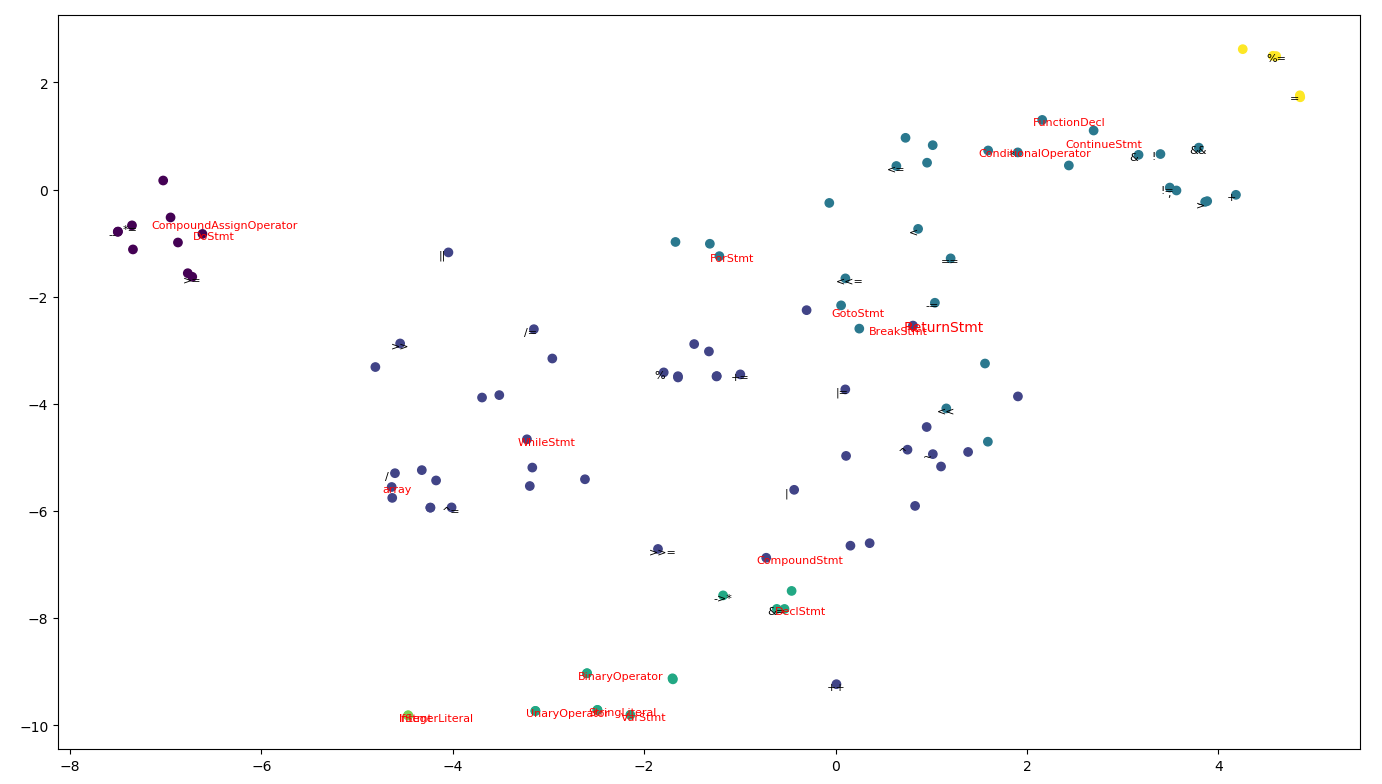}
    	\caption{A t-SNE plot of the learned node representations, where different node colors denote different clusters.}
    	\label{fig:nodeRepresentationVisualization}
\end{figure}

Based on the characteristics of GGANN model, the feature representation of each node is computed by learning its local structure, i.e. its neighbor nodes and the edges among them. Thus, the nodes with similar local structure should have similar representation. In order to evaluate whether the GGANN model can learn this structure similarity, we use K-mean, an unsupervised clustering method, to cluster the learned node feature vectors, and visualize the clustering results in two-dimensional space through t-SNE. The experimental results are shown in Figure~\ref{fig:nodeRepresentationVisualization}, where it can be observed that the node representation vectors learned from the GGANN model reflect certain clustering phenomenon.

In order to further analyze whether nodes in a cluster share similar semantic information, we compare the corresponding semantic information of nodes in each cluster. A few clusters and the corresponding nodes are shown in Table~\ref{table:clustering}, where it can be observed that the GGANN model does learn the semantic characteristics of nodes. For example, ForStmt, ContinueStmt, GotoStmt, BreakStmt, and etc are clustered into cluster 3, because most of them are related to control flow. Similarly, UnaryOperator, BinaryOperator, StringLiteral, IntegerLiteral, and etc, are clustered to cluster 1, because these operators are basically related to data reference/manipulation, while the statement >= and -=, as well as CompoundAssignOperator, are clustered to cluster 2, because most of them are related to composite operations.

\begin{table*}[htbp]
    \centering
    \caption{The partial results of Kmeans clustering ($K=5$).}
  \setlength{\tabcolsep}{0.7mm}{
    \begin{tabular}{clp|p{0.1em} }
    \hline
    Cluster	&Symbols\\
    \hline
   1	&varStmt,StringLiteral,UnaryOperator,BinaryOperator,IntegerLiteral \\
  2	&DoStmt,CompoundAssignOperator,$\gg$=,+=,-= \\
  3	&ConditionalOperator,FunctionDel,ContinueStmt,\&,$||$,!=,==,$>$=,ForStmt,GotoStmt,BreakStmt,ReturnStmt\\
  4&	WhileStmt, CompoundStmt\\
  5&\%= $,$ =\\
    \hline
    \end{tabular}}\label{table:clustering}
    \end{table*}


\subsubsection{Attention Analysis}

\begin{figure}[htb]
    	\centering
    	\includegraphics [width=0.5\textwidth]{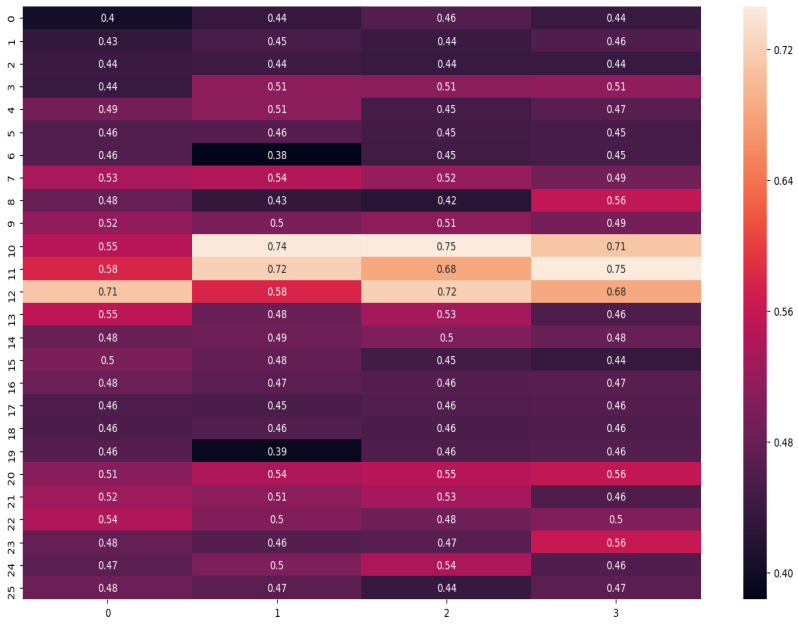}
    	\caption{ The attention scores of nodes when merged
to graph representation.
}
    	\label{fig:attentionHotMap}
\end{figure}

GGANN defines a graph-level aggregation function for graph-level output, as shown in Formula~\eqref{GGANN_eq_4}. Each node has a weight (i.e., attention) calculated through attention mechanism, which reflects the similarity between the node and the whole FDA graph. In order to verify whether the node attention learned by GGANN indeed reflects the node importance, we present the attention of all nodes in an FDA graph in the form of thermodynamic chart, as shown in Figure~\ref{fig:attentionHotMap}, which has 104 rectangular bars, each of which corresponds to a node of the FDA graph. The lighter the color, the higher the node weight. In Figure~\ref{fig:attentionHotMap}, the nodes with higher weights are BinaryOperator, CallExpr, ImplicitCastExpr, IfStmt, DeclStmt, and CompundStmt, with weights being 0.71, 0.74, 0.75, 0.75, 0.72 and 0.74, respectively. These nodes are basically located at the higher levels of AST, which are generally more abstract than the nodes at lower levels and contain more information about the whole program. The experiment results reflect that the GGANN model indeed learned the node importance and paid more attention to the important nodes.

\section{Conclusion and Future Work}\label{conclusion}
To address the problem of the limited social interactions among the users of the online programming service, we propose to use source-code mining technique to promote social interactions. As the first attempt to achieve the goal, in this work, we propose a graph-based program classification model, which can classify a give program with high accuracy by understanding the structural and semantic information of the program. More specifically, in this work, the FDA graph is proposed to convert a program's source codes to an intermediate representation to be fed into the graph network models, and a graph network model GGANN is proposed for the program classification application. As far as we know, this is the first work that applies the graph network models to the program classification application. The experimental evaluation illustrates that the proposed work successfully learns the grammatical structure and semantic information of various programming tasks, including the tasks with similar descriptions, and can generalize well to the test set, comparing with the state-of-art program classification work, TBCNN. In the future, we will further mine the correlation implicated in the source code through the integration of the compiling techniques, accelerate the training speed of the GGANN model through the sampling and pooling techniques, and extend the GGANN model to other programming languages and other tasks.


%

\section*{Acknowledgment}
This work was partially supported by the National Science Foundation of China, Project Nos. 61232001,61173169, 91646115, and 60903222; the Science Foundation of Hunan, Project No. 2016JJ2149 and No.018JJ3012; and the Major Science and Technology Research Program for Strategic Emerging Industry of Hunan, Grant No. 2012GK4054.

\ifCLASSOPTIONcaptionsoff
  \newpage
\fi



\bibliographystyle{IEEEtran}
\bibliography{MyRef}

\begin{thebibliography}{10}
\providecommand{\url}[1]{#1}
\csname url@samestyle\endcsname
\providecommand{\newblock}{\relax}
\providecommand{\bibinfo}[2]{#2}
\providecommand{\BIBentrySTDinterwordspacing}{\spaceskip=0pt\relax}
\providecommand{\BIBentryALTinterwordstretchfactor}{4}
\providecommand{\BIBentryALTinterwordspacing}{\spaceskip=\fontdimen2\font plus
\BIBentryALTinterwordstretchfactor\fontdimen3\font minus
  \fontdimen4\font\relax}
\providecommand{\BIBforeignlanguage}[2]{{%
\expandafter\ifx\csname l@#1\endcsname\relax
\typeout{** WARNING: IEEEtran.bst: No hyphenation pattern has been}%
\typeout{** loaded for the language `#1'. Using the pattern for}%
\typeout{** the default language instead.}%
\else
\language=\csname l@#1\endcsname
\fi
#2}}
\providecommand{\BIBdecl}{\relax}
\BIBdecl

\bibitem{b00}
\BIBentryALTinterwordspacing
Github. [Online]. Available: \url{https://github.com}
\BIBentrySTDinterwordspacing

\bibitem{b01}
\BIBentryALTinterwordspacing
Topcoder. [Online]. Available: \url{https://www.topcoder.com}
\BIBentrySTDinterwordspacing

\bibitem{b02}
\BIBentryALTinterwordspacing
Educoder. [Online]. Available: \url{https://www.educoder.net}
\BIBentrySTDinterwordspacing

\bibitem{b0}
A.~Barab\'asi, \emph{The Formula: The Universal Laws of Success}.\hskip 1em
  plus 0.5em minus 0.4em\relax Little Brown and Company, 2018.

\bibitem{b1}
M.~Allamanis, E.~T. Barr, P.~Devanbu, and S.~Charles, ``A survey of machine
  learning for big code and naturalness,'' \emph{Journal ACM Computing
  Surveys}, vol.~51, no.~4, 2018.

\bibitem{b36}
M.~Khajah, R.~Lindsey, and M.~Mozer, ``How deep is knowledge tracing?''
  \emph{Educational Data Mining}, pp. 94--101, 2016.

\bibitem{b3}
M.~Allamanis, H.~Peng, and C.~Sutton, ``A convolutional attention network for
  extreme summarization of source code,'' \emph{arXiv preprint
  arXiv:1602.03001}, 2016.

\bibitem{b8}
A.~Hindle, E.~Barr, Z.~Su, P.~Devanbu, and M.~Gabel, ``On the naturalness of
  software,'' in \emph{International Conference on Software Engineering}, 2012.

\bibitem{b10}
X.~Gu, H.~Zhang, D.~Zhang, and S.~Kim, ``Deep {API} learning,'' in \emph{ACM
  Sigsoft International Symposium on Foundations of Software Engineering},
  2016, pp. 631--642.

\bibitem{b16}
Y.~Lu, G.~Li, R.~Miao, and Z.~Jin, ``Learning embeddings of {API} tokens to
  facilitate deep learning based program processing,'' \emph{Lehner F., Fteimi
  N. (eds) Knowledge Science, Engineering and Management. KSEM 2016. Lecture
  Notes in Computer Science}, vol. 9983, 2016.

\bibitem{b29}
V.~Raychev, M.~Vechev, and E.~Yahav, ``Code completion with statistical
  language models,'' \emph{{ACM} Sigplan Notices}, vol.~49, no.~6, pp.
  419--428, 2014.

\bibitem{b5}
N.~Bui, L.~Jiang, and Y.~Yu, ``Cross-language learning for program
  classification using bilateral tree-based convolutional neural networks,'' in
  \emph{AAAI}, 2018.

\bibitem{b20}
C.~Maddison and D.~Tarlow, ``Structured generative models of natural source
  code,'' \emph{CoRR}, vol. abs/1401.0514, 2014.

\bibitem{b24}
H.~Peng, L.~Mou, G.~Li, Y.~Liu, L.~Zhang, and Z.~Jin, ``Building program vector
  representations for deep learning,'' \emph{In: Zhang S., Wirsing M., Zhang Z.
  (eds) Knowledge Science, Engineering and Management. KSEM 2015. Lecture Notes
  in Computer Science}, vol. 9403, 2015.

\bibitem{b25}
L.~Mou, G.~Li, L.~Zhang, T.~Wang, and Z.~Jin, ``Convolutional neural network
  over tree structures for programming language processing,'' in \emph{AAAI},
  2016, pp. 1287--1293.

\bibitem{b26}
L.~Mou, R.~Men, G.~Li, Y.~Xu, L.~Zhang, R.~Yan, and Z.~Jin, ``Recognizing
  entailment and contradiction by tree-based convolution,'' \emph{CoRR}, vol.
  abs/1512.08422, 2015.

\bibitem{b38}
V.~Raychev, M.~Vechev, and A.~Krause, ``Predicting program properties from "big
  code",'' in \emph{{ACM} Sigplan-sigact Symposium on Principles of Programming
  Languages}, 2015.

\bibitem{b2}
M.~Allamanis, M.~Brockschmidt, and M.~Khademi, ``Learning to represent programs
  with graphs,'' in \emph{International Conference on Learning
  Representations}, 2017.

\bibitem{b13}
Y.~Li, D.~Tarlow, M.~Brockschmidt, and R.~Zemel, ``Gated graph sequence neural
  networks,'' in \emph{International Conference on Learning Representations},
  2015.

\bibitem{b4}
A.~Bhoopchand, T.~Rocktschel, E.~Barr, and S.~Riedel, ``Learning python code
  suggestion with a sparse pointer network,'' in \emph{International Conference
  on Learning Representations}, 2017.

\bibitem{b7}
M.~Defferrard, X.~Bresson, and P.~Vandergheynst, ``Convolutional neural
  networks on graphs with fast localized spectral filtering,'' in
  \emph{Advances in Neural Information Processing Systems}, 2016, pp.
  3844--3852.

\bibitem{b17}
T.~Kipf and M.~Welling, ``Semi-supervised classification with graph
  convolutional networks,'' in \emph{International Conference on Learning
  Representations}, 2017.

\bibitem{b21}
M.~Gori, G.~Monfardini, and F.~Scarselli, ``A new model for learning in graph
  domains,'' in \emph{Proceedings of 2005 IEEE International Joint Conference
  on Neural Networks, 2005.}, vol.~2, 2005, pp. 729--734.

\bibitem{b30}
S.~Franco, G.~Marco, T.~A. Chung, H.~Markus, and M.~Gabriele, ``The graph
  neural network model,'' \emph{IEEE Transactions on Neural Networks}, vol.~20,
  no.~1, pp. 61--80, 2009.

\bibitem{b22}
F.~Miller, A.~Vandome, and J.~Mcbrewster, \emph{Abstract syntax tree}.\hskip
  1em plus 0.5em minus 0.4em\relax Alphascript Publishing, 2010.

\bibitem{b34}
X.~Liu and Y.~Tang, ``Similarity analysis of malware's function-call graphs,''
  \emph{Computer Engineering \& Science}, 2014.

\bibitem{b39}
B.~M. Olenick, C.~A. Szyperski, D.~G. Hunt, G.~L. Hughes, W.~A. Manis, and
  T.~Zmrhal, ``Accessing and manipulating data in a data flow graph,'' 2010.

\bibitem{b18}
D.~Kingma and J.~Ba, ``Adam: A method for stochastic optimization,''
  \emph{Computer Science}, 2014.

\bibitem{b33}
X.~Glorot and Y.~Bengio, ``Understanding the difficulty of training deep
  feedforward neural networks,'' in \emph{Proceedings of the the Thirteenth
  International Conference on Artificial Intelligence and Statistics}, pp.
  249--256.

\bibitem{b31}
N.~Srivastava, G.~Hinton, A.~Krizhevsky, I.~Sutskever, and R.~Salakhutdinov,
  ``Dropout: A simple way to prevent neural networks from overfitting,''
  \emph{Journal of Machine Learning Research}, vol.~15, no.~1, pp. 1929--1958,
  2014.

\end{thebibliography}
%

%

\begin{IEEEbiography}{Mingming Lu} is current an associate professor in the School of Computer Science and Engineering, Central South University, China. He received his PhD in Computer Science from Florida Atlantic University, USA in 2008. His current research areas include deep learning, source code mining, and edge computing. He has published over 60 journal and conference papers. 
\end{IEEEbiography}

\begin{IEEEbiographynophoto}{Dingwu Tan} is current a graduate student in the School of Computer Science and Engineering, Central South University, China. His current research areas include deep learning and source code mining.
\end{IEEEbiographynophoto}


\begin{IEEEbiographynophoto}{Zailiang Chen} is current an associate professor in the School of Computer Science and Engineering, Central South University, China. His current research areas include deep learning, image processing, and source code mining.
\end{IEEEbiographynophoto}

\begin{IEEEbiographynophoto}{Naixue Xiong} is current a Professor in College of Intelligence and Computing, Tianjin University, China. He received his both PhD degrees in Wuhan University (about sensor system engineering), and Japan Advanced Institute of Science and Technology (about dependable sensor networks), respectively. Before he attended Tianjin University, he worked in Northeastern State University, Georgia State University, Wentworth Technology Institution, and Colorado Technical University (full professor about 5 years) about 10 years. His research interests include Cloud Computing, Security and Dependability, Parallel and Distributed Computing, Networks, and Optimization Theory. 

Dr. Xiong published over 300 international journal papers and over 100 international conference papers. Some of his works were published in IEEE JSAC, IEEE or ACM transactions, ACM Sigcomm workshop, IEEE INFOCOM, ICDCS, and IPDPS. He has been a General Chair, Program Chair, Publicity Chair, PC member and OC member of over 100 international conferences, and as a reviewer of about 100 international journals, including IEEE JSAC, IEEE SMC (Park: A/B/C), IEEE Transactions on Communications, IEEE Transactions on Mobile Computing, IEEE Trans. on Parallel and Distributed Systems. He is serving as an Editor-in-Chief, Associate editor or Editor member for over 10 international journals (including Associate Editor for IEEE Tran. on Systems, Man \& Cybernetics: Systems, Associate Editor for Information Science, Editor-in-Chief for Journal of Internet Technology (JIT), and Editor-in-Chief for Journal of Parallel \& Cloud Computing (PCC)), and a guest editor for over 10 international journals, including Sensor Journal, WINET and MONET. He has received the Best Paper Award in the 10th IEEE International Conference on High Performance Computing and Communications (HPCC-08) and the Best student Paper Award in the 28th North American Fuzzy Information Processing Society Annual Conference (NAFIPS2009). 

  Dr. Xiong is the Chair of “Trusted Cloud Computing” Task Force, IEEE Computational Intelligence Society (CIS),
http://www.cs.gsu.edu/~cscnxx/index-TF.html, and the Industry System Applications Technical Committee, http://ieee-cis.org/technical/isatc/; He is a Senior member of IEEE Computer Society, E-mail: xiongnaixue@gmail.com. 
\end{IEEEbiographynophoto}

\begin{IEEEbiographynophoto}{Haifeng Li} is current a proforssor in the School of Geosciences and Info-Physics, Central South University,
Changsha, China. His current research areas include deep learning, geo data mining, and source code mining.
\end{IEEEbiographynophoto}




\end{document}